\documentclass[Proceedings,letterpaper]{ascelike-new}
\usepackage[utf8]{inputenc}
\usepackage[T1]{fontenc}
\usepackage{lmodern}
\usepackage{graphicx}
\usepackage{amsmath}
\usepackage{newtxtext,newtxmath}
\usepackage[colorlinks=true,citecolor=black,linkcolor=black]{hyperref}

\NewDocumentCommand{\evalat}{sO{\big}mm}{%
  \IfBooleanTF{#1}
   {\mleft. #3 \mright|_{#4}}
   {#3#2|_{#4}}%
}

\begin{document}
\title{SIFT-Aided Rectified 2D-DIC for Displacement and Strain Measurements in Asphalt Concrete Testing}

\author[1*]{Zehui Zhu}
\author[2]{Imad L. Al-Qadi}

\affil[1]{Department of Civil and Environmental Engineering, University of Illinois Urbana-Champaign. Email: zehuiz2@illinois.edu}
\affil[2]{Department of Civil and Environmental Engineering, University of Illinois Urbana-Champaign. Email: alqadi@illinois.edu}
\affil[*]{Corresponding Author}

\footnote{This paper was accepted for publication in the Journal of Transportation Engineering, Part B: Pavements. DOI: 10.1061/JPEODX.PVENG-1401.}

\maketitle

\begin{abstract}
Two-dimensional digital image correlation (2D-DIC) is a widely used optical technique to measure displacement and strain during asphalt concrete (AC) testing.  An accurate 2-D DIC measurement can only be achieved when the camera's principal axis is perpendicular to the planar specimen surface. However, this requirement may not be met during testing due to device constraints. This paper proposes a simple and reliable method to correct errors induced by non-perpendicularity. The method is based on image feature matching and rectification. No additional equipment is needed. A theoretical error analysis was conducted to quantify the effect of a non-perpendicular camera alignment on measurement accuracy. The proposed method was validated numerically using synthetic images and experimentally in an AC fracture test. It achieved relatively high accuracy, even under considerable camera rotation angle and large deformation. As a pre-processing technique, the proposed method showed promising performance in assisting the recently developed CrackPropNet for automated crack propagation measurement under a non-perpendicular camera alignment.
\end{abstract}

\section{Introduction}
\label{sec:intro}
Digital image correlation (DIC) is an optical displacement and strain measurement technique. Since its introduction for asphalt concrete (AC) application, DIC has become a vital tool in evaluating AC material properties, validating test protocols, and verifying theoretical models \cite{chehab2007viscoelastoplastic,birgisson2008determination,birgisson2009optical,safavizadeh2017utilizing,rivera2021illinois}. The most popular DIC algorithms include the two-dimensional (2D)-DIC and stereo-DIC. The 2D-DIC uses a single fixed camera to measure the in-plane displacement of nominal planar objects. However, the 2D-DIC method is not applicable when the test specimen surface is non-planar, or the out-of-plane displacement is non-negligible. On the other hand, the stereo-DIC method, which uses two synchronized cameras, was proposed to measure 3D displacement based on the binocular stereo-vision principle \cite{pan2009two}. 

The 2D-DIC remains one of the most popular optical technologies because of the following advantages: 
\begin{enumerate}
    \item \emph{Simple experimental setup}: For 2D-DIC, only a single charge-coupled device (CCD) camera is needed. In contrast, the stereo-DIC requires two synchronized cameras \cite{pan2018digital}.
    \item \emph{Simple algorithm}: The computation of the displacement field in 2D-DIC solely relies on the correlation between deformed images captured after deformation and a reference image captured before deformation. On the other hand, stereo-DIC entails a complex procedure that includes stereo-calibration of the two-camera unit, stereo matching (i.e., cross-camera matching), temporal matching, and triangulation for reconstructing 3D coordinates \cite{lin2022path}. Hence, the stereo-DIC has more uncertainty due to the different algorithms chosen in each step, thereby impacting the ultimate precision of measurements \cite{balcaen2017stereo,zhong2019comparative}.
\end{enumerate}

However, an accurate 2-D DIC measurement can only be achieved when the following requirements are met \cite{sutton2009image}:
\begin{enumerate}
    \item Specimen surface is planar.
    \item Out-of-plane deformation of the specimen is small enough.
    \item Imaging system does not suffer from geometric distortion.
    \item Charged-couple device sensor is parallel to the specimen surface (i.e., the camera's principal axis is perpendicular to the specimen surface).
\end{enumerate}

\begin{figure}[h!]
    \centering
    \includegraphics[trim=0 0 0 0,clip,width=0.8\textwidth]{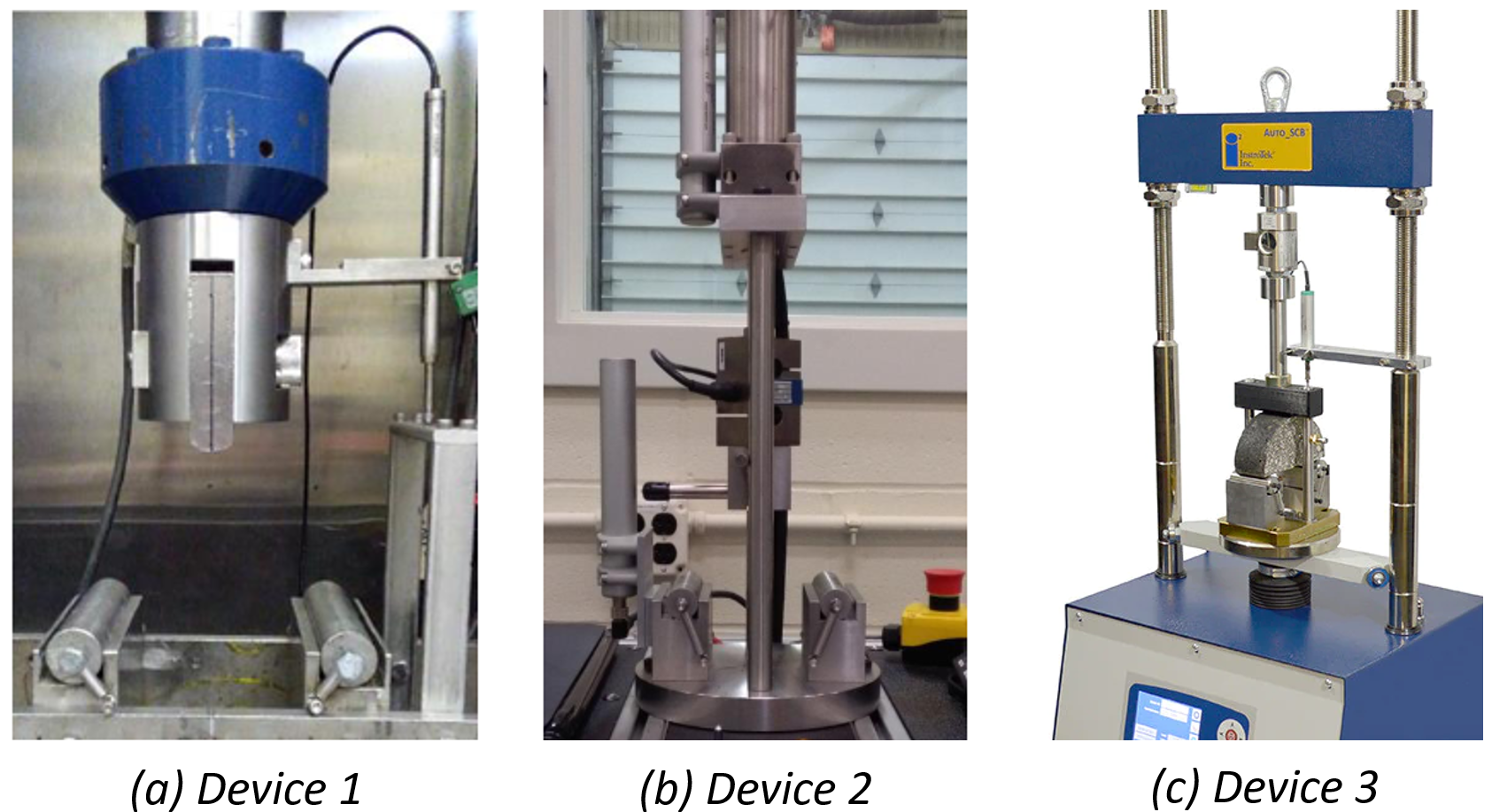}
    \caption{I-FIT testing devices.}
    \label{fig:test_devices}
\end{figure}

The first three requirements can be easily satisfied in most AC laboratory tests. However, equipment limitations often make the last requirement difficult to meet. For example, as shown in Fig.\ref{fig:test_devices}, only one of the three testing devices for the Illinois-flexibility index test (I-FIT) offers open space to satisfy the perpendicularity requirement \cite{ozer2017evaluation}. 

In the past decade, researchers have proposed a few methods to alleviate the effect of non-perpendicularity on 2D-DIC. They can be categorized into two groups based on their fundamental principles. The first is modifying experimental setup or transforming images. For example, \citeN{pan2013high} proposed changes to the imaging system. A bilateral telecentric lens is recommended for its relatively high sensitivity to in-plane displacement and insensitivity to out-of-plane displacement. Moreover, \citeN{pan2013high1} suggested attaching a non-deformable compensation specimen onto the testing surface. During loading, the compensation specimen moves rigidly with the testing specimen. The measurement error can be corrected by applying a parametric model fitted using the compensation specimen's displacement field. The second, quite differently, applies rectification to transform images collected under a non-perpendicular camera setting \cite{lava2011error}. The standard camera calibration procedure was followed using a regular grid pattern. The camera's extrinsic and intrinsic parameters were determined and used to conduct the rectification. DIC analysis was conducted on the rectified images, which can be collected under a perpendicular camera setting.

While the mentioned approaches effectively correct errors caused by non-perpendicularity, the simplicity of 2D-DIC is compromised by the requirement of additional equipment, such as a special lens, compensation specimen, or grid pattern. To fully leverage the benefits of 2D-DIC, this paper proposes a simple and reliable method with no additional equipment needed. The proposed method was validated numerically using synthetic images and experimentally in an I-FIT. In addition, as a pre-processing technique, the proposed method was applied to assist CrackPropNet in measuring crack propagation on an I-FIT specimen surface.


\section{Two-Dimensional DIC Principle}
\begin{figure}[ht!]
    \centering
    \includegraphics[trim=0 0 0 0,clip,width=0.55\textwidth]{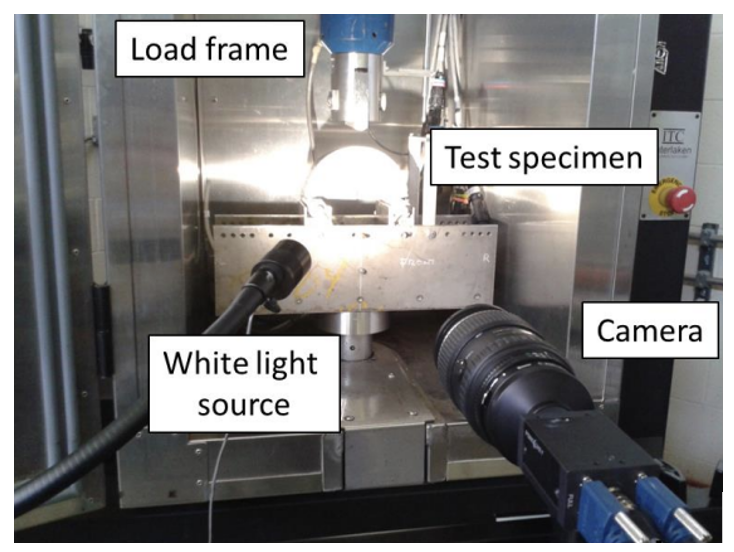}
    \caption{2-D DIC setup.}
    \label{fig:dic_set_up}
\end{figure}

This section reviews the fundamentals of 2D-DIC. Fig.\ref{fig:dic_set_up} shows a 2-D DIC system comprises of a camera, a light source, and a computer.

\begin{figure}[ht!]
    \centering
    \includegraphics[trim=0 0 0 0,clip,width=0.95\textwidth]{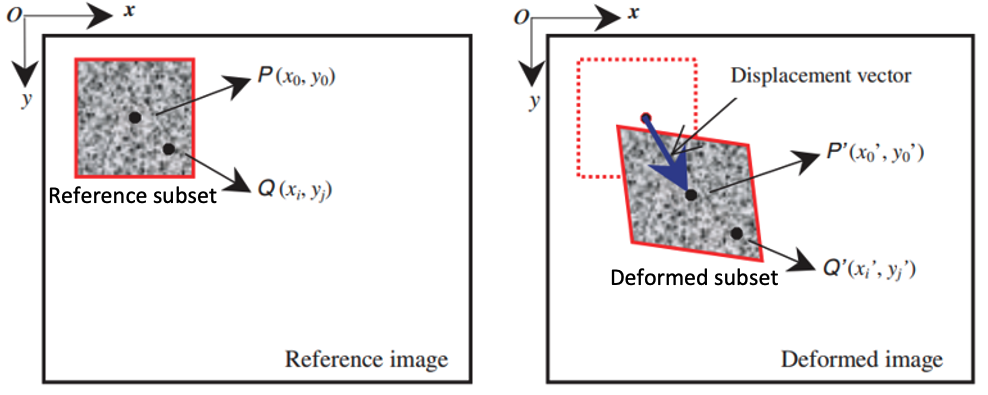}
    \caption{Establish correspondences between a reference and a deformed images \protect\cite{pan2009two}.}
    \label{fig:abm}
\end{figure}

The DIC works by tracking pixel movement in a series of images. It is achieved by establishing gray value correspondences. For example, as shown in Fig.\ref{fig:abm}, a match between a square reference subset and a deformed subset is established. The matching is attained by searching for a correlation coefficient extremum. Equation \ref{eqn:correlation} lists commonly used correlation criteria \cite{pan2009two}. Compared to cross-correlation (CC) and sum-of-squared differences (SSD), the zero-normalized cross-correlation (ZNCC) and zero-normalized sum-of-squared differences (ZNSSD) offer better performance against noise. They are less insensitive to lighting fluctuations (e.g., offset and linear scale).

\begin{equation}
    \begin{aligned}
    CC &: \sum_{i=-M}^{M} \sum_{j=-M}^{M} f(x_i,y_j) g(x_i',y_j')\\
    ZNCC &: \frac{\sum_{i=-M}^{M} \sum_{j=-M}^{M}[f(x_i,y_j)-f_m]\times[g(x_i',y_j')-g_m]}{\sqrt{\sum_{i=-M}^{M} \sum_{j=-M}^{M} [f(x_i,y_j)-f_m]^2}\sqrt{\sum_{i=-M}^{M} \sum_{j=-M}^{M} [g(x_i',y_j')-g_m]^2}} \\
    SSD &: \sum_{i=-M}^{M} \sum_{j=-M}^{M} [f(x_i,y_j) - g(x_i',y_j')]^2 \\
    ZNSSD &: \sum_{i=-M}^{M} \sum_{j=-M}^{M} [\frac{f(x_i,y_j)-f_m}{\sqrt{\sum_{i=-M}^{M} \sum_{j=-M}^{M} [f(x_i,y_j)-f_m]^2}} - \frac{g(x_i',y_j')-g_m}{\sqrt{\sum_{i=-M}^{M} \sum_{j=-M}^{M} [g(x_i',y_j')-g_m]^2}}]^2
\end{aligned}
\label{eqn:correlation}
\end{equation}

where $f(x_i,y_j)$ is gray value at $(x_i, y_j)$ in the reference subset; $g(x_i',y_j')$ is gray value at $(x_i', y_j')$ in the deformed subset; $f_m$ and $g_m$ are mean gray values of the reference and deformed subset, respectively.

Once the matching is established, the in-plane displacement vector of point $P$ can be calculated based on coordinate difference. Subsequently, displacement of other points in the same subset (e.g., $Q$) can be computed using shape functions $\gamma$ and $\delta$ in Eq.\ref{eqn:shape_fn}. A commonly used first-order shape function enables translation, shear, rotation, and combinations.

\begin{equation}
    \begin{cases}
    x_i' = x_i+\gamma(x_i, y_i)\\
    y_j' = y_j+\delta(x_i, y_i)
\end{cases}
\label{eqn:shape_fn}
\end{equation}

The above only discussed calculating the displacement vector for a single point. The reliability-guided (RG)-DIC algorithm, adopted by open-source software Ncorr, is often used to attain full-field measurement \cite{pan2009two,blaber2015ncorr}. First, the algorithm finds a reasonable initial guess of the displacement vector ($\mathbf{p}_0=(\gamma(x, y), \delta(x, y))$) for a seed point. The seed point's correlation coefficient is also calculated. Second, the displacement vectors of the four neighboring points of the seed point are calculated using $\mathbf{p}_0$ as their initial guess. Their correlation coefficients are inserted into a max heap binary tree. Third, the tree's root with the highest correlation coefficient is popped, and its $\mathbf{p}$ is used as the initial guess to calculate displacement vectors of the uncomputed neighboring points. The third step is repeated until all points in the region of interest (ROI) are calculated.


\section{Research Objective}
This paper addresses 2D-DIC measurement errors caused by non-perpendicularity. 

A theoretical analysis was conducted to understand the displacement error quantitatively. A simple and reliable method was proposed to correct such errors. The proposed method was validated numerically using synthetic images and experimentally in an I-FIT. Moreover, the proposed method was applied as a pre-processing technique to assist CrackPropNet in measuring crack propagation on an I-FIT specimen surface.


\section{Theoretical Error Analysis of A Non-Perpendicular Camera Alignment}
Perpendicularity is crucial in using 2D-DIC. If the camera's principal axis is not perpendicular to the planar specimen surface, measurement errors are unavoidable. This section presents a theoretical study of the significant factors affecting these errors. 

\begin{figure}[ht!]
    \centering
    \includegraphics[trim=0 0 0 0,clip,width=0.45\textwidth]{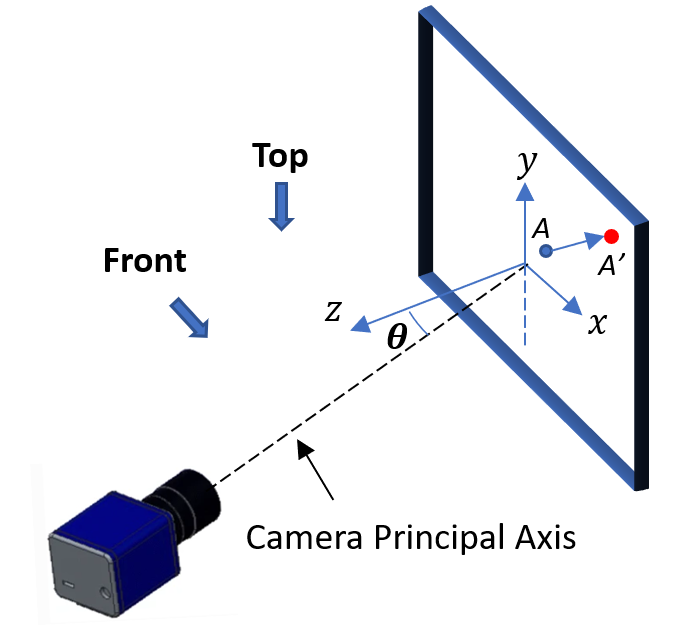}
    \caption{Camera's principal axis is not perpendicular to the planar specimen surface.}
    \label{fig:non_perpendicular}
\end{figure}

As shown in Fig.\ref{fig:non_perpendicular}, a pinhole camera's principal axis is rotated $\theta$ with respect to the $y$-axis. Displacement error was assessed by assuming a point $A(x_A,y_A)$ moved to $A'(x_A+\Delta x, y_A+\Delta y)$.  

\begin{figure}[ht!]
    \centering
    \includegraphics[trim=0 0 0 0,clip,width=0.95\textwidth]{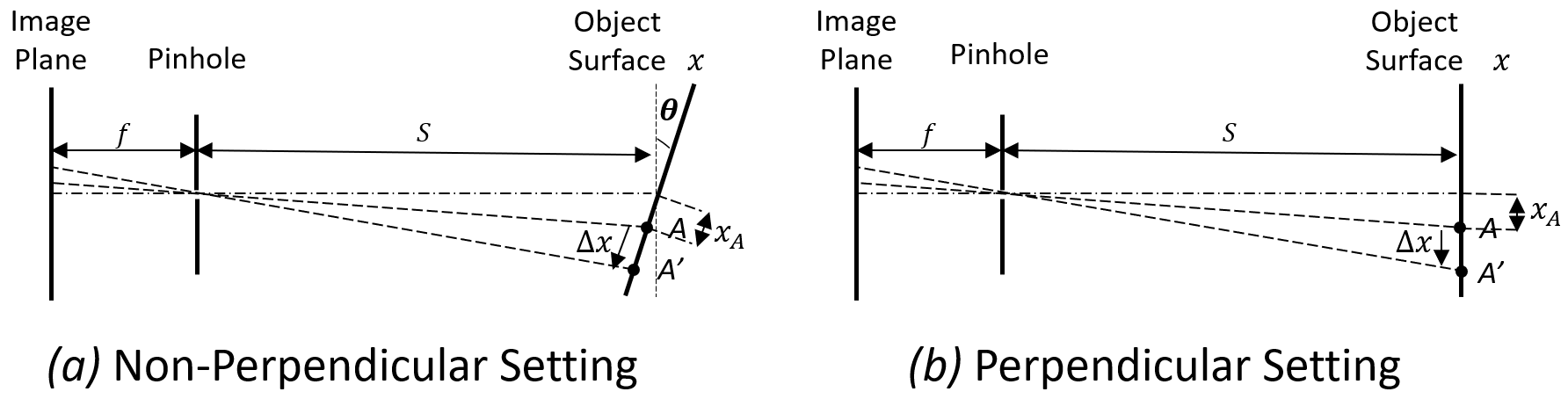}
    \caption{Top view of the pinhole camera model under non-perpendicular and perpendicular camera settings.}
    \label{fig:theory_x}
\end{figure}

Fig.\ref{fig:theory_x} shows the top view of Fig.\ref{fig:non_perpendicular}. $f$ refers to the focal length, the distance between the pinhole plane and the image plane. $S$ is the distance between the pinhole plane and the planar object surface. 

Under perpendicular setting (Fig.\ref{fig:theory_x}\emph{(b)}), projected displacement $\Delta x$ on the image plane is presented as follows:

\begin{equation}
    \Delta x' = \frac{\Delta x}{S}f 
\label{eqn:x_per}
\end{equation}

Meanwhile, under non-perpendicular setting (Fig.\ref{fig:theory_x}\emph{(a)}), by applying the concept of the similar triangle, projected displacement on the image plane was derived:

\begin{equation}
    \Delta x'' = f[\frac{(x_A + \Delta x)\cos \theta}{S-(x_A + \Delta x)\sin \theta} - \frac{x_A\cos \theta}{S-x_A\sin \theta}] 
\label{eqn:x_nonper}
\end{equation}

\begin{figure}[ht!]
    \centering
    \includegraphics[trim=0 0 0 0,clip,width=0.95\textwidth]{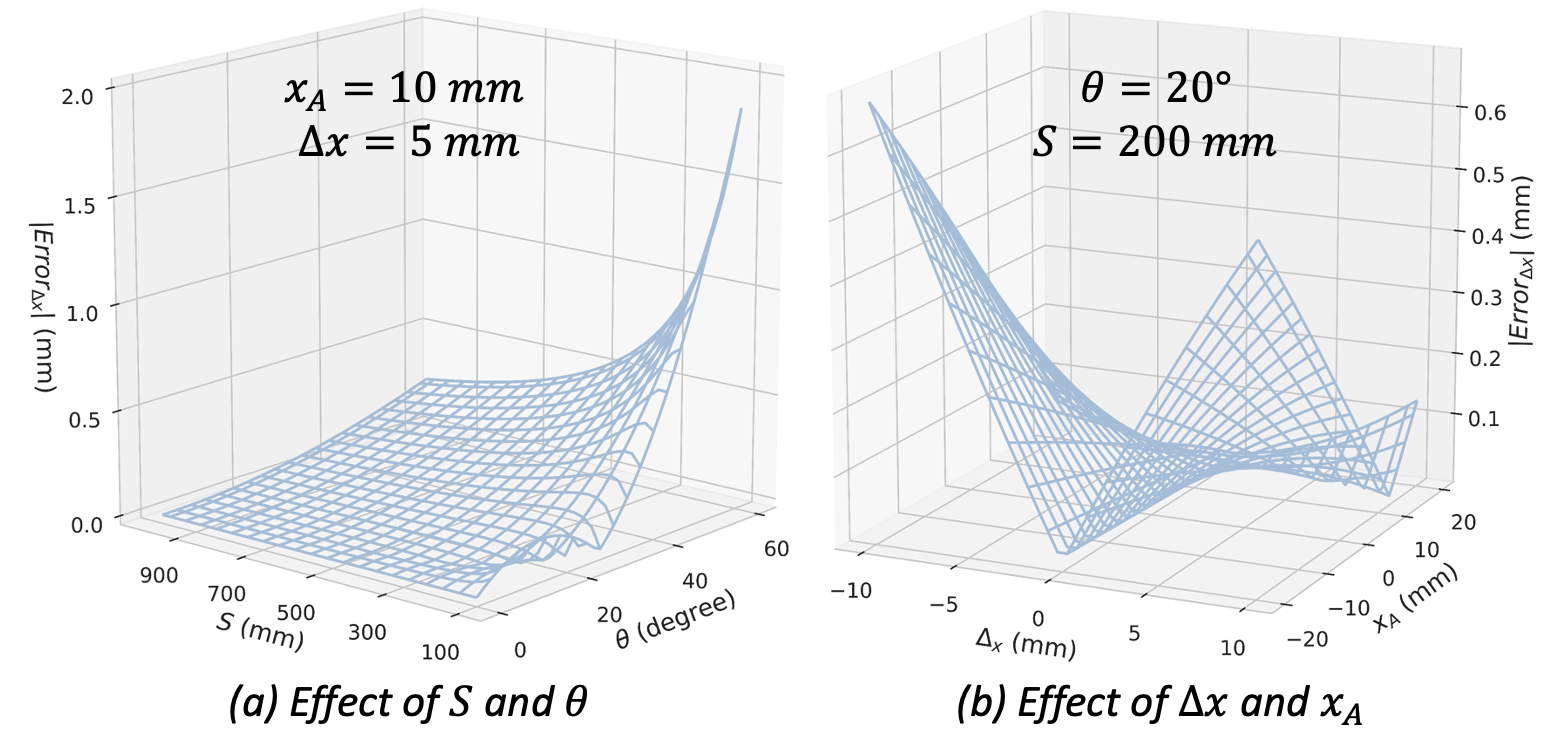}
    \caption{Effect of $\Delta x$, $S$, $x_A$, and $\theta$ on $|Error_{\Delta x}|$.}
    \label{fig:error_u}
\end{figure}

Eq.\ref{eqn:x_error} shows the absolute error of projected displacement $\Delta x$ on the image plane. The magnitude of error depends on $f$, $S$, $\theta$, $\Delta x$, and $x_A$. The error increases as the focal length ($f$) increases. Fig.\ref{fig:error_u} illustrates the effect of the remaining parameters. The parametric values considered here are typical in AC laboratory tests. Due to the high-dimensional nature of Eq.\ref{eqn:x_error}, Fig.\ref{fig:error_u} illustrates partial dependencies by fixing other parameters. The error increases exponentially with a larger camera rotation angle ($\theta$) and a smaller pinhole-object distance ($S$). Moreover, a more significant displacement $\Delta_x$ causes a larger error. However, the relationship between $x_A$ and $|Error_{\Delta x}|$ is non-monotonic.

\begin{equation}
    |Error_{\Delta x}| = f|\frac{\Delta x}{S}-\frac{(x_A + \Delta x)\cos \theta}{S-(x_A + \Delta x)\sin \theta}+\frac{x_A\cos \theta}{S-x_A\sin \theta}| 
\label{eqn:x_error}
\end{equation}

\begin{figure}[ht!]
    \centering
    \includegraphics[trim=0 0 0 0,clip,width=0.95\textwidth]{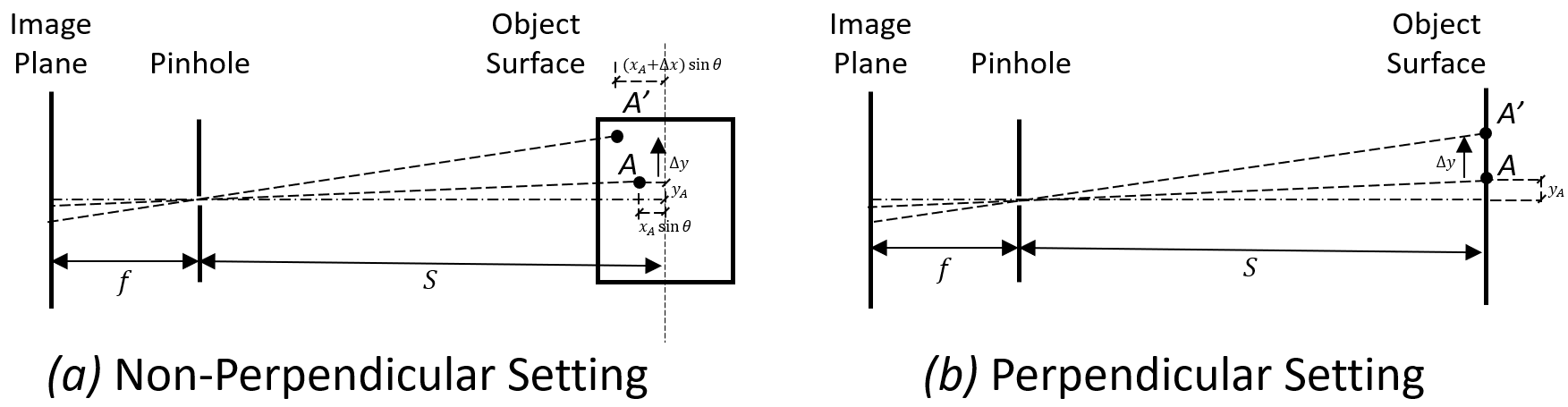}
    \caption{Front view of the pinhole camera model under non-perpendicular and perpendicular camera settings.}
    \label{fig:theory_y}
\end{figure}

Fig.\ref{fig:theory_y} shows the front view of Fig.\ref{fig:non_perpendicular}. 

Under perpendicular setting (Fig.\ref{fig:theory_y}\emph{(b)}), the projected displacement $\Delta y$ on the image plane is given by Eq.\ref{eqn:y_per}:

\begin{equation}
    \Delta y' = \frac{\Delta y}{S}f 
\label{eqn:y_per}
\end{equation}

Meanwhile, under the non-perpendicular setting (Fig.\ref{fig:theory_y}\emph{(a)}), the projected displacement $\Delta y$ on the image plane can be obtained using the concept of similar triangles:

\begin{equation}
    \Delta y'' = f(\frac{y_A+\Delta y}{S-(x_A+\Delta x)\sin \theta}-\frac{y_A}{S-x_A \sin \theta}) 
\label{eqn:y_nonper}
\end{equation}

\begin{figure}[ht!]
    \centering
    \includegraphics[trim=0 0 0 0,clip,width=0.8\textwidth]{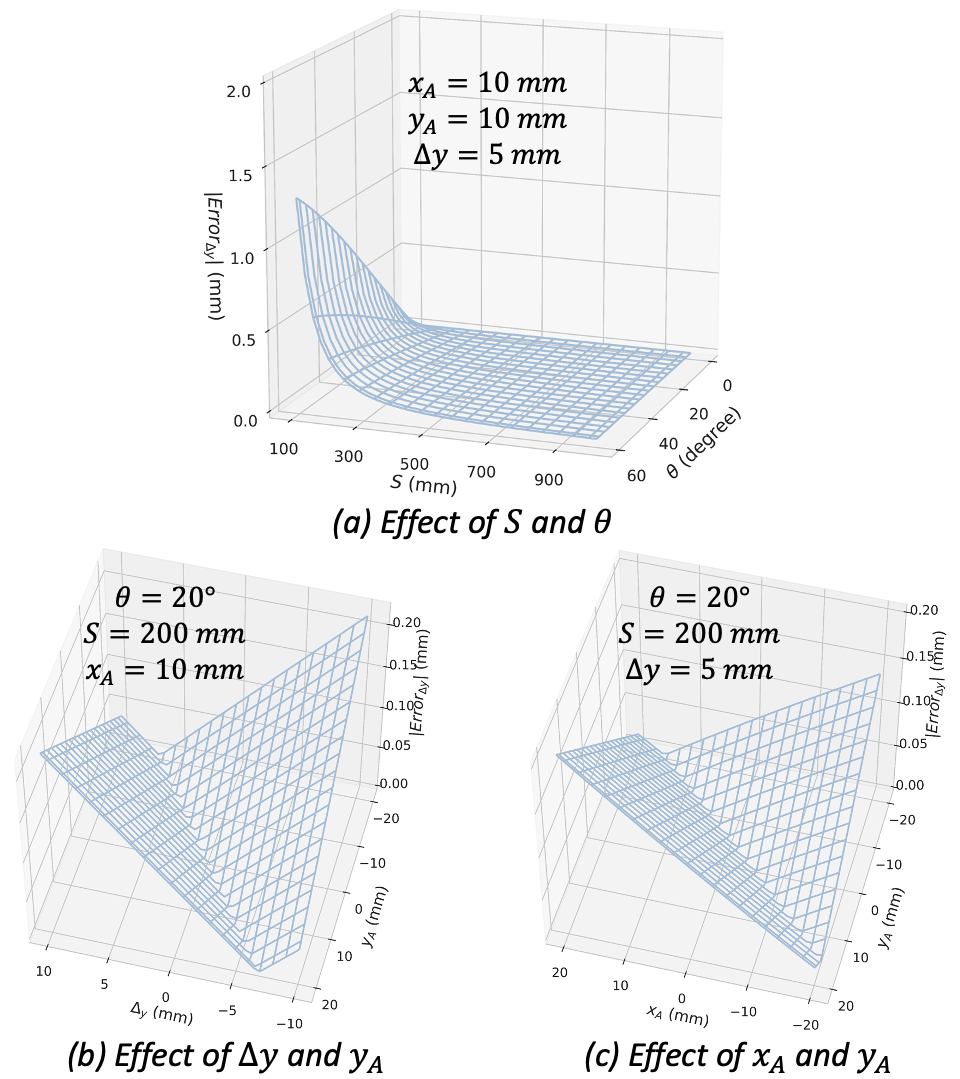}
    \caption{Effect of $\Delta y$, $S$, $y_A$, $x_A$, $\Delta x$, and $\theta$ on $|Error_{\Delta y}|$.}
    \label{fig:error_v}
\end{figure}

Eq.\ref{eqn:y_error} presents the absolute error of projected displacement $\Delta y$ on the image plane, which depends on $f$, $S$, $\theta$, $\Delta x$, $\Delta y$, $x_A$, and $y_A$. The error increases with increasing focal length ($f$) and decreasing $\Delta x$. Fig. \ref{fig:error_v} demonstrates the impact of the remaining parameters, highlighting partial dependencies while keeping other parameters fixed. The error generally increases with a larger camera rotation angle ($\theta$) and a smaller pinhole-object distance ($S$). However, the effect of $x_A$, $y_A$, and $\Delta y$ is not monotonic.

\begin{equation}
    |Error_{\Delta y}| = f|\frac{\Delta y}{S} -\frac{y_A+\Delta y}{S-(x_A+\Delta x)\sin \theta}+\frac{y_A}{S-x_A \sin \theta}| 
\label{eqn:y_error}
\end{equation}

\section{SIFT-Aided Rectified 2D-DIC}
This paper proposes a new method to compensate for measurement errors induced by non-perpendicular camera settings. The method, namely, scale-invariant feature transform (SIFT)-aided rectified 2D-DIC, consists of six steps:
\begin{enumerate}
    \item Image acquisition.
    \item Key point extraction.
    \item Key point matching.
    \item Homography matrix estimation.
    \item Image rectification.
    \item 2D-DIC analysis.
\end{enumerate}

This section discusses the details of each step. An example is also presented for further explanation.

\subsection{Image Acquisition}
\begin{figure}[ht!]
    \centering
    \includegraphics[trim=0 0 0 0,clip,width=0.7\textwidth]{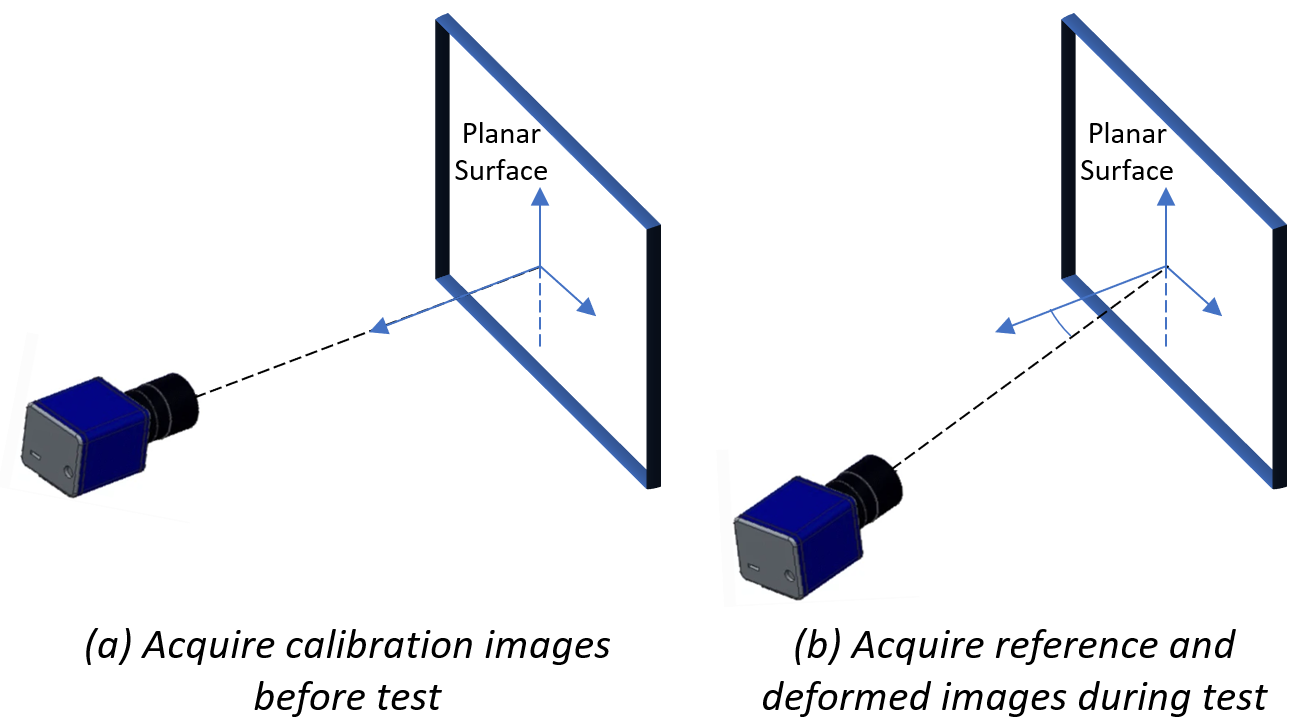}
    \caption{Image acquisition.}
    \label{fig:img_acq}
\end{figure}

When a non-perpendicular camera alignment is unavoidable in a 2D-DIC measurement, several calibration images of the specimen are collected before testing in a perpendicular setting, as shown in Fig.\ref{fig:img_acq}\emph{(a)}. Next, reference and deformed images are acquired using the same camera under the non-perpendicular setting during the test, as shown in Fig.\ref{fig:img_acq}\emph{(b)}. It is recommended not to change the camera settings (e.g., aperture, ISO, shutter speed) during this procedure. Changing camera settings would require intrinsic camera calibration. The camera settings must be carefully tuned such that all acquired images are in focus.

\subsection{Key Point Extraction}
\label{sec:featureExtraction}
\begin{figure}[ht!]
    \centering
    \includegraphics[trim=0 0 0 0,clip,width=0.85\textwidth]{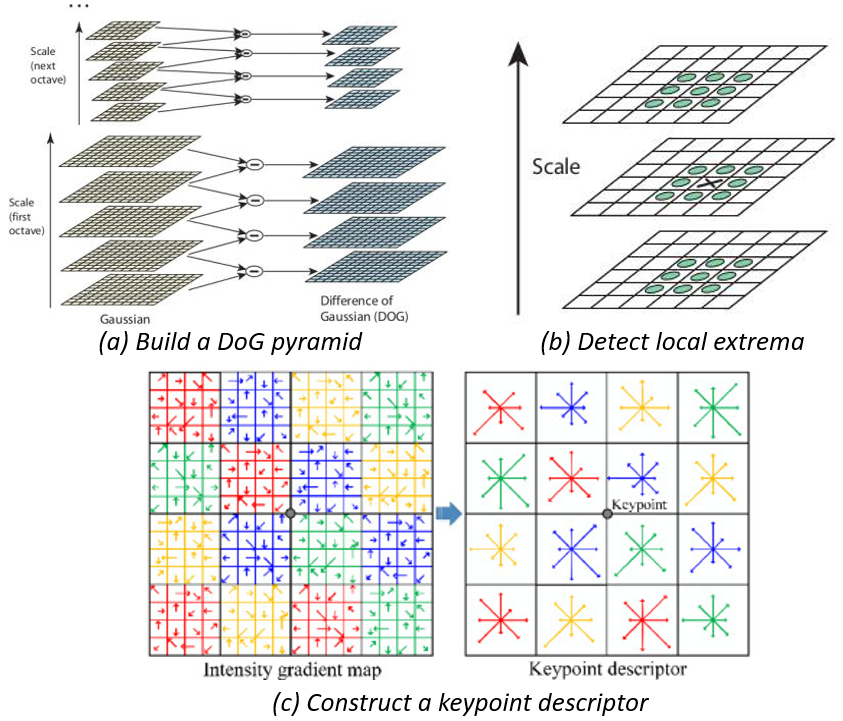}
    \caption{Schematic of SIFT algorithm.}
    \label{fig:sift}
\end{figure}

Key points and their descriptors are extracted from the calibration and the reference images using the SIFT algorithm \cite{lowe2004distinctive}:
\begin{itemize}
    \item \emph{Build a difference-of-Gaussian (DoG) pyramid.} Image ($I(x,y)$) is repeatedly convolved with 2D Gaussian filters ($G(x,y,\sigma_i)$) with increasing $\sigma_i$, creating Gaussian images ($L(x,y,\sigma_i)$) in Fig.\ref{fig:sift}(\emph{a}). DoG images are produced by subtracting adjacent Gaussian images. (Eq.\ref{eqn:dog}).

    
    \begin{equation}
        D(x,y,\sigma_i)=L(x,y,\sigma_i)-L(x,y,\sigma_{i-1})
    \label{eqn:dog}
    \end{equation}
    
    \item \emph{Detect key points.} As shown in Fig.\ref{fig:sift}(\emph{b}), a key point is located by comparing a pixel with its neighbors in a $3\times3\times3$ cubic. Local maxima or minima are considered key points.
    \item \emph{Refine key point locations into the sub-pixel level.} Eq.\ref{eqn:taylor} shows the Taylor expansion of $D(x,y,\sigma)$ up to the quadratic terms \cite{brown2002invariant}. Assume $\mathbf{x_0}$ is a key point location found in the previous step. By taking the first derivative of $D$ and setting it to zero, offset $\mathbf{h}$ is obtained (Eq.\ref{eqn:offset}). $\mathbf{x_0}+\mathbf{h}$ is the refined location of the key point.

    \begin{equation}
        D(\mathbf{x_0}+\mathbf{h}) \approx D(\mathbf{x_0})+\evalat{(\frac{\partial D}{\partial \mathbf{x}})^T}{\mathbf{x}=\mathbf{x_0}}\mathbf{h}+\frac{1}{2}\mathbf{h}^T\frac{\partial^2 D}{\partial^2 \mathbf{x}}\mathbf{h}
    \label{eqn:taylor}
    \end{equation}

    \begin{equation}
        \mathbf{h}=-(\frac{\partial^2 D}{\partial^2 \mathbf{x}})^{-1}(\frac{\partial D}{\partial \mathbf{x}})^T
    \label{eqn:offset}
    \end{equation}
    
    \item \emph{Assign a dominant orientation to each key point.} Each key point is described concerning its dominant orientation to achieve rotation invariance. A histogram of intensity gradient orientations is created for each key point's neighborhood, using 36 bins with 10$^\circ$ coverage per bin. The peak orientation in the histogram is considered the dominant orientation and assigned to the key point.
    \item \emph{Construct key point descriptors.} As shown in Fig.\ref{fig:sift}\emph{(c)}, a 16x16 neighborhood around each key point is used to construct its descriptor. The gradient direction and magnitude are computed at each pixel. Next, the magnitude is weighted by a Gaussian window based on the pixel's distance to the key point. Then, the neighborhood is divided into sixteen 4$\times$4 sub-regions. For each sub-region, an eight-bin orientation histogram is created (Fig.\ref{fig:sift}\emph{(c)}). Hence, a vector of size 128 is formed as the descriptor of each key point.
\end{itemize}

\begin{figure}[ht!]
    \centering
    \includegraphics[trim=0 0 0 0,clip,width=0.8\textwidth]{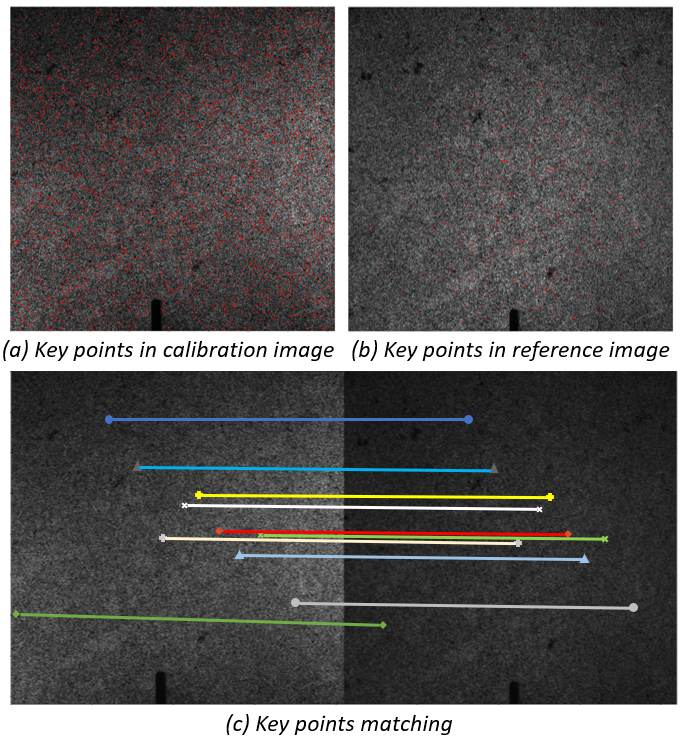}
    \caption{Key points are matched between  calibration and reference images.}
    \label{fig:matching}
\end{figure}

For example, as shown in Fig.\ref{fig:matching}\emph{(a)} and \emph{(b)}, the above-described procedure was followed to extract key points and their descriptors on the calibration and the reference image. The images were collected from an I-FIT specimen surface using a Point Grey Gazelle 4.1MP Mono camera with a resolution of $2048 \times 2048$. The calibration and the reference images identified 152,150 and 24,151 key points, respectively. Only one percent of the key points were marked to provide better visualization.

\subsection{Key Point Matching}
\label{sec:featureMatching}
A one-to-one key point match is established by finding the shortest euclidean distance between the 128-sized descriptor vectors. To filter out false matches, if the second-shortest distance is close to the shortest distance (i.e., $\frac{d_{closest}}{d_{next\; closest}} \geq \delta$), the match will be discarded. The threshold ($\delta$) is user-defined and typically ranges from 0.3 to 0.8.

For example, 1,946 key point matches were identified between the calibration and the reference image. A $\delta$ of 0.55 was used because it eliminated most false matches while preserving enough correct matches. Fig.\ref{fig:matching}\emph{(c)} only shows 0.5\% of the matches for better visualization purposes.

\subsection{Homography Matrix Estimation}
\label{sec:homography}
\begin{figure}[ht!]
    \centering
    \includegraphics[trim=0 0 0 0,clip,width=0.8\textwidth]{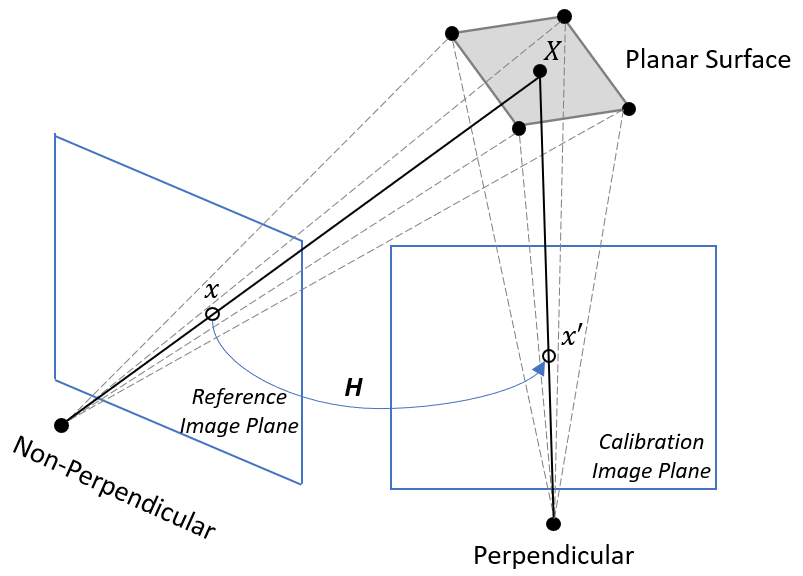}
    \caption{Illustration of homography.}
    \label{fig:homography}
\end{figure}

Under the pinhole camera model, a homography matrix $H$ relates any two images of the same planar surface. For example, because the I-FIT specimen surface is planar, the calibration image is related to the reference image by $\mathbf{H}$. In Fig.\ref{fig:homography}, $\mathbf{x}([x,y,1]^T)$ refers to the projected coordinates of point $X$ in the reference coordinate system, and $\mathbf{x}'([x',y',1]^T)$ represents the projected coordinates of point $X$ in the calibration coordinate system. The relationship between $\mathbf{x}$ and $\mathbf{x}'$ is described by $\mathbf{H}$ (Eq. \ref{eqn:xhx}).

\begin{equation}
    \lambda \mathbf{x}' = \mathbf{H}\mathbf{x}
\label{eqn:xhx}
\end{equation}

where $\lambda$ is a scaling factor; $\mathbf{H}$ has eight degrees of freedom, as shown in Eq.\ref{eqn:H}.

\begin{equation}
\mathbf{H} = \begin{bmatrix}h_{11} & h_{12} & h_{13}\\h_{21} & h_{22} & h_{23}\\h_{31} & h_{32} & 1\end{bmatrix}
\label{eqn:H}
\end{equation}

Each key point match provides two linearly independent equations, as shown in Eq.\ref{eqn:matching}. Thus, to estimate $\mathbf{H}$, at least four matches are needed. 

\begin{equation}
\begin{aligned}
\lambda \begin{bmatrix}x_i'\\y_i'\\1\end{bmatrix} = \begin{bmatrix}h_{11} & h_{12} & h_{13}\\h_{21} & h_{22} & h_{23}\\h_{31} & h_{32} & 1\end{bmatrix}\begin{bmatrix}x_i\\y_i\\1\end{bmatrix}
\\ \Rightarrow \begin{bmatrix}x_i'\\y_i'\\1\end{bmatrix} \times \begin{bmatrix}\mathbf{h_1}^T\mathbf{x}_i\\\mathbf{h_2}^T\mathbf{x}_i\\\mathbf{h_3}^T\mathbf{x}_i\end{bmatrix} = \begin{bmatrix} y_i'\mathbf{h_3}^T\mathbf{x}_i-\mathbf{h_2}^T\mathbf{x}_i\\\mathbf{h_1}^T\mathbf{x}_i-x_i'\mathbf{h_3}^T\mathbf{x}_i\\x_i'\mathbf{h_2}^T\mathbf{x}_i-y_i'\mathbf{h_1}^T\mathbf{x}_i \end{bmatrix}
\\ \Rightarrow \begin{bmatrix}\mathbf{0}^T & -\mathbf{x}_i^T & y_i'\mathbf{x}_i^T\\\mathbf{x}_i^T & \mathbf{0}^T & -x_i'\mathbf{x}_i^T\\-y_i'\mathbf{x}_i^T & x_i'\mathbf{x}_i^T & \mathbf{0}^T\end{bmatrix}\begin{bmatrix}\mathbf{h_1}\\\mathbf{h_2}\\\mathbf{h_3}\end{bmatrix}=\mathbf{0}
\end{aligned}
\label{eqn:matching}
\end{equation}

Given the matching key points, the random sample consensus (RANSAC) algorithm was used to estimate the homography matrix as below \cite{fischler1981random}:
\begin{enumerate}
    \item Randomly sample four key point matches.
    \item Solve for the eight unknown parameters in $\mathbf{H}$.
    \item Determine number of matches that fit the solved $\hat{\mathbf{H}}$ according to Eq.\ref{eqn:inlier}. $\epsilon$ is often set between 1 to 10 pixels. An $\epsilon$ of 5 was used in this paper.
    
    \begin{equation}
        ||x'-\hat{\mathbf{H}}x||_2 \leq \epsilon
    \label{eqn:inlier}
    \end{equation}

    \item Steps 1 to 3 are repeated 2000 times. Each time, an estimated $\hat{\mathbf{H}}$ is rejected if there are too few fitting matches or is kept if the consensus set is larger than the previously saved solution.
\end{enumerate}

The RANSAC algorithm was chosen because of its robustness against noise and outliers. To enhance the reliability of the estimated homography matrix, it is suggested to repeat the above procedure on all calibration images and take the mean. For example, steps \ref{sec:featureExtraction} through \ref{sec:homography} were repeated for each of the five calibration images, and the mean estimated homography matrix follows:



\begin{equation}
\hat{\bar{\mathbf{H}}} = \begin{bmatrix}1.0640e+00 & 6.4001e-04 & -2.1895e+02\\-1.4929e-02 & 9.3542e-01 & 2.7591e+01\\-2.6070e-05 & 4.2958e-07 & 1.0000e+00\end{bmatrix}
\label{eqn:estimatedH}
\end{equation}

\subsection{Image Rectification}
\label{sec:rectification}
The reference and deformed images collected under a non-perpendicular camera setting are rectified according to Eq.\ref{eqn:rectify}. Because calculated coordinates on the right-hand side may not be integers, a bilinear interpolation was used. 

\begin{equation}
    R(x,y) = I(\frac{\hat{\bar{\mathbf{H}}}_{11}x+\hat{\bar{\mathbf{H}}}_{12}y+\hat{\bar{\mathbf{H}}}_{13}}{\hat{\bar{\mathbf{H}}}_{31}x+\hat{\bar{\mathbf{H}}}_{32}y+\hat{\bar{\mathbf{H}}}_{33}}, \frac{\hat{\bar{\mathbf{H}}}_{21}x+\hat{\bar{\mathbf{H}}}_{22}y+\hat{\bar{\mathbf{H}}}_{23}}{\hat{\bar{\mathbf{H}}}_{31}x+\hat{\bar{\mathbf{H}}}_{32}y+\hat{\bar{\mathbf{H}}}_{33}})
\label{eqn:rectify}
\end{equation}

where $R(x,y)$ is gray value at $(x, y)$ in the rectified image; $\hat{\bar{\mathbf{H}}}$ is estimated homography matrix; $I$ refers to raw image being rectified.

\subsection{2D-DIC Analysis}
DIC analysis should be conducted on rectified images as they can be considered acquired under the perpendicular setting.

\begin{figure}[ht!]
    \centering
    \includegraphics[trim=0 0 0 0,clip,width=0.95\textwidth]{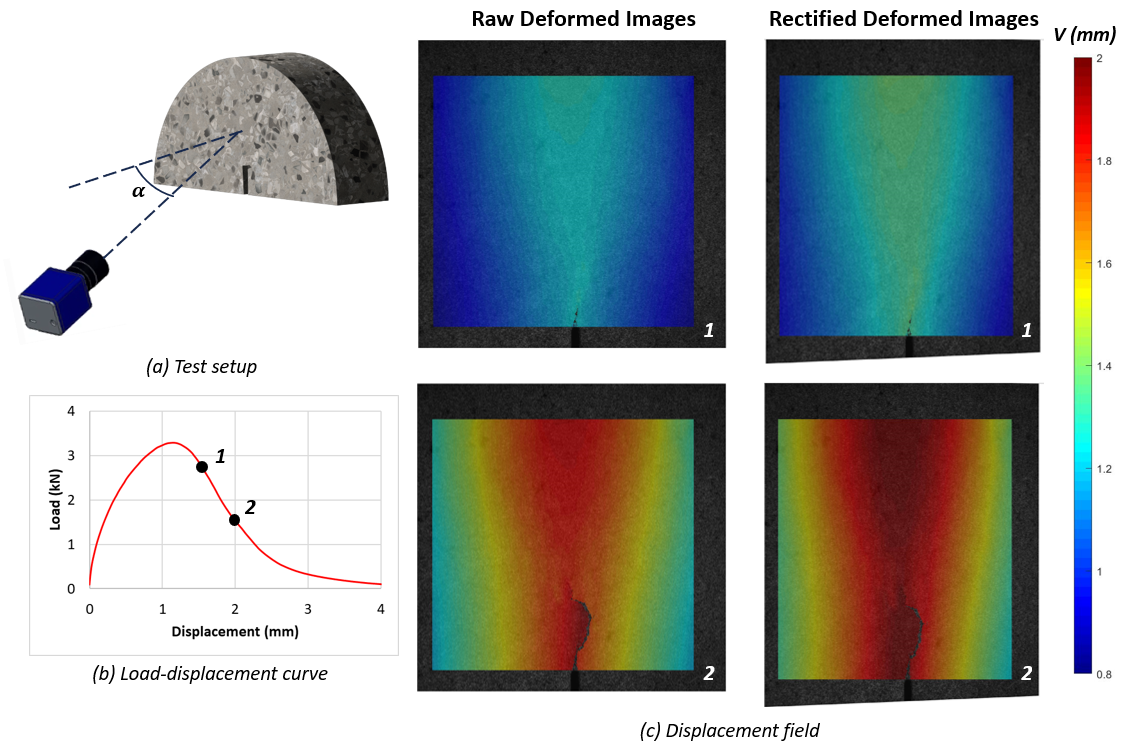}
    \caption{DIC-measured vertical displacement field ($\mathbf{v}$) on raw and rectified deformed images.}
    \label{fig:dic_example}
\end{figure}

Fig.\ref{fig:dic_example}(\emph{c}) shows the DIC-measured vertical displacement field ($\mathbf{v}$) using raw and rectified deformed images. The raw images were collected under a non-perpendicular camera setting while conducting  I-FIT, as shown in Fig.\ref{fig:dic_example}(\emph{a}). The I-FIT is a fracture test that has been widely used to characterize the cracking potential of AC. More information can be found elsewhere \cite{ozer2016development,ozer2016fracture,zhu2020quantification}. Significant discrepancies existed in the measured displacement fields. The accuracy will be discussed in the next section.

\section{Method Validation}
The proposed method was validated numerically using synthetic images and experimentally on an I-FIT. Numerical validation offers precise control and reliability due to the known displacement field. Experimental validation is crucial to verify real-world applicability.

\subsection{Numerical Validation}
\subsubsection{Synthetic Images}
First, a series of static images were collected on an I-FIT specimen surface using a Point Grey Gazelle 4.1MP Mono camera with a resolution of $2048 \times 2048$. The imaging system was carefully set up to ensure the camera's principal axis was perpendicular to the specimen surface. 

\begin{figure}[ht!]
    \centering
    \includegraphics[trim=0 0 0 0,clip,width=0.5\textwidth]{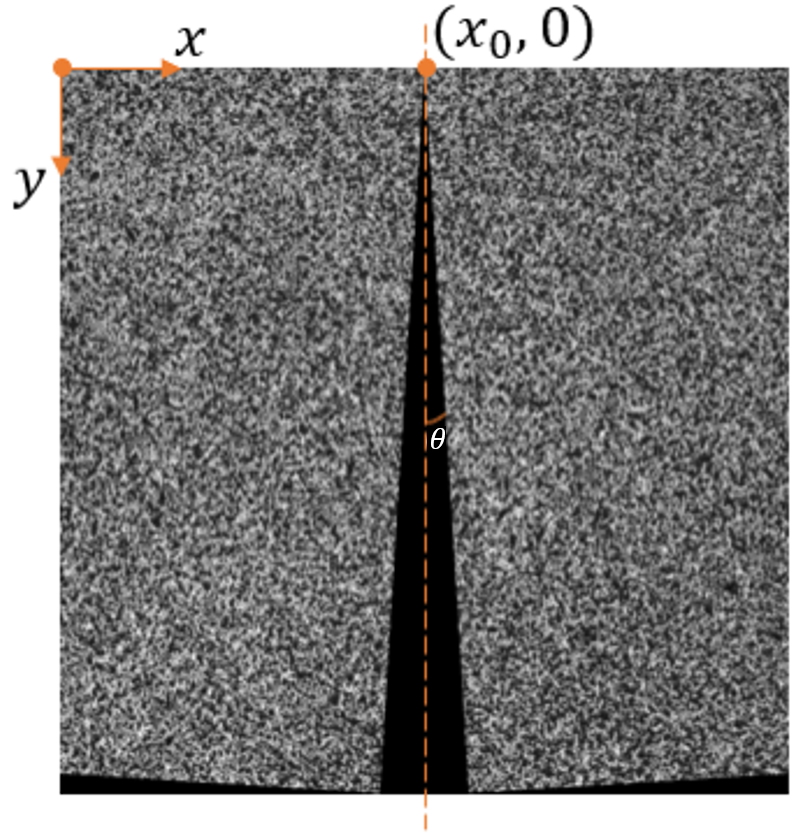}
    \caption{Displacement fields used to generate deformed images.}
    \label{fig:displ}
\end{figure}

\begin{figure}[ht!]
    \centering
    \includegraphics[trim=0 0 0 0,clip,width=0.98\textwidth]{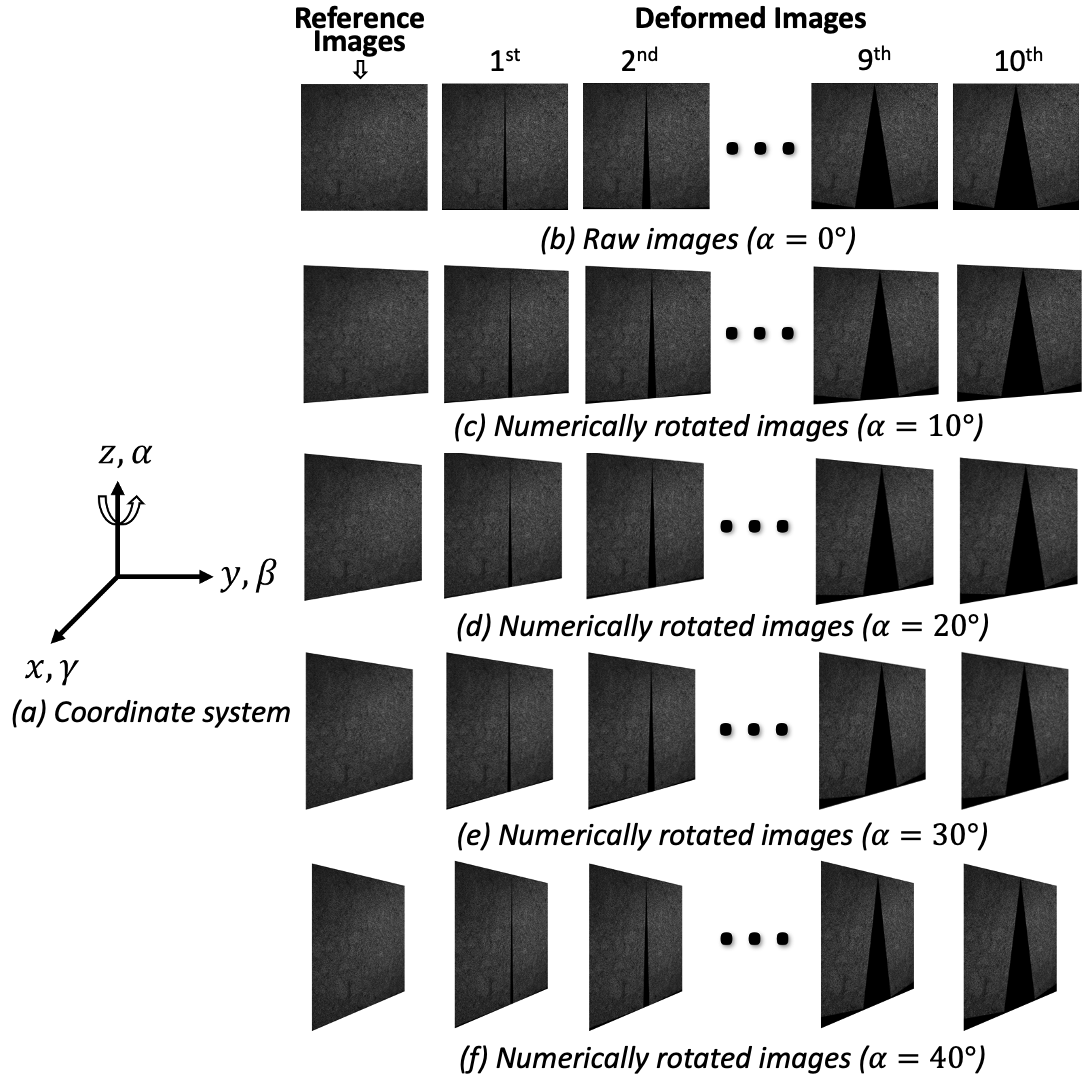}
    \caption{Euler rotation schema and synthetic images.}
    \label{fig:rotation}
\end{figure}

Second, the displacement field shown in Fig.\ref{fig:displ} was used to generate deformed images. It involves rotation concerning point $(x_0,0)$ only. The displacement vector $(u, v)$ at each pixel can be calculated using Eq.\ref{eqn:displ}. Counterclockwise rotation was defined as positive \cite{zhu2022}. As shown in Fig.\ref{fig:rotation}\emph{(b)}, ten deformed images were generated with $\theta = {1^\circ, 2^\circ, \dots, 10^\circ}$.

\begin{equation}
\begin{cases}
    u = \sin (\theta+\tan^{-1} (\frac{x-x_0}{y}))\sqrt{(x-x_0)^2+y^2} + x_0 - x \\
    v = \cos (\theta+\tan^{-1} (\frac{x-x_0}{y}))\sqrt{(x-x_0)^2+y^2} - y
\end{cases}
\label{eqn:displ}
\end{equation}

Third, the reference and synthetic deformed images were numerically rotated to simulate non-perpendicular camera settings. Only rotation with respect to $z$-axis was considered (Fig.\ref{fig:rotation}\emph{(a)}). Four rotating angles $\alpha = {10^\circ, 20^\circ, 30^\circ, 40^\circ}$ were used, resulting in four series of images, as shown in Fig.\ref{fig:rotation}\emph{(c-f)}. The rotation matrix was given by Eq.\ref{eqn:rotation} \cite{goldstein2002classical}.


\begin{equation}
\begin{aligned}
\mathbf{R} 
&= \mathbf{R}_z(\alpha)\mathbf{R}_y(\beta)\mathbf{R}_x(\gamma) \\
&= \begin{bmatrix} \cos\alpha & -\sin\alpha & 0 \\ \sin\alpha & \cos\alpha & 0 \\ 0 & 0 & 1 \end{bmatrix}\begin{bmatrix} \cos\beta & 0 & \sin\beta \\ 0 & 1 & 0 \\ -\sin\beta & 0 & \cos\beta \end{bmatrix}\begin{bmatrix} 1 & 0 & 0 \\ 0 & \cos\gamma & -\sin\gamma \\ 0 & \sin\gamma & \cos\gamma \end{bmatrix} \\
&= \begin{bmatrix} \cos\alpha\cos\beta & \cos\alpha\sin\beta\sin\gamma-\sin\alpha\cos\gamma & \cos\alpha\sin\beta\cos\gamma+\sin\alpha\sin\gamma \\ \sin\alpha\cos\beta & \sin\alpha\sin\beta\sin\gamma+\cos\alpha\cos\gamma & \sin\alpha\sin\beta\cos\gamma-\cos\alpha\sin\gamma \\ -\sin\beta & \cos\beta\sin\gamma & \cos\beta\cos\gamma \end{bmatrix} 
\end{aligned}
\label{eqn:rotation}
\end{equation}

where $\alpha$, $\beta$, and $\gamma$ are Euler angles around $z$-, $y$-, and $x$-axis, respectively (Fig.\ref{fig:rotation}\emph{(a)}). 

\subsubsection{Experimental Setup}
First, DIC analysis was conducted on raw image series (Fig.\ref{fig:rotation}\emph{(b)}) acquired under a perpendicular camera setting. The results were referred to as \emph{'Perpendicular'} in Figs.\ref{fig:num_vis} and \ref{fig:res1}.

Second, two analyses were conducted on each series of numerically rotated images(Fig.\ref{fig:rotation}\emph{(c-f)}), which simulate non-perpendicular camera settings. 
\begin{itemize}
    \item DIC analysis was conducted directly on numerically rotated images (referred to as \emph{'Non-Perpendicular'} in Figs.\ref{fig:num_vis} and \ref{fig:res1}).
    \item SIFT-aided rectified 2D-DIC was conducted. The numerically rotated images were first rectified before undergoing the DIC analysis (referred to as \emph{'Rectified Non-Perpendicular'} in Figs.\ref{fig:num_vis} and \ref{fig:res1}). The raw reference image was used as the calibration image.
\end{itemize}

All analyses were on the same ROI. A subset size of $23 \times 23$ was chosen through an iterative process.

\subsubsection{Results}
\begin{figure}[ht!]
    \centering
    \includegraphics[trim=0 0 0 0,clip,width=0.98\textwidth]{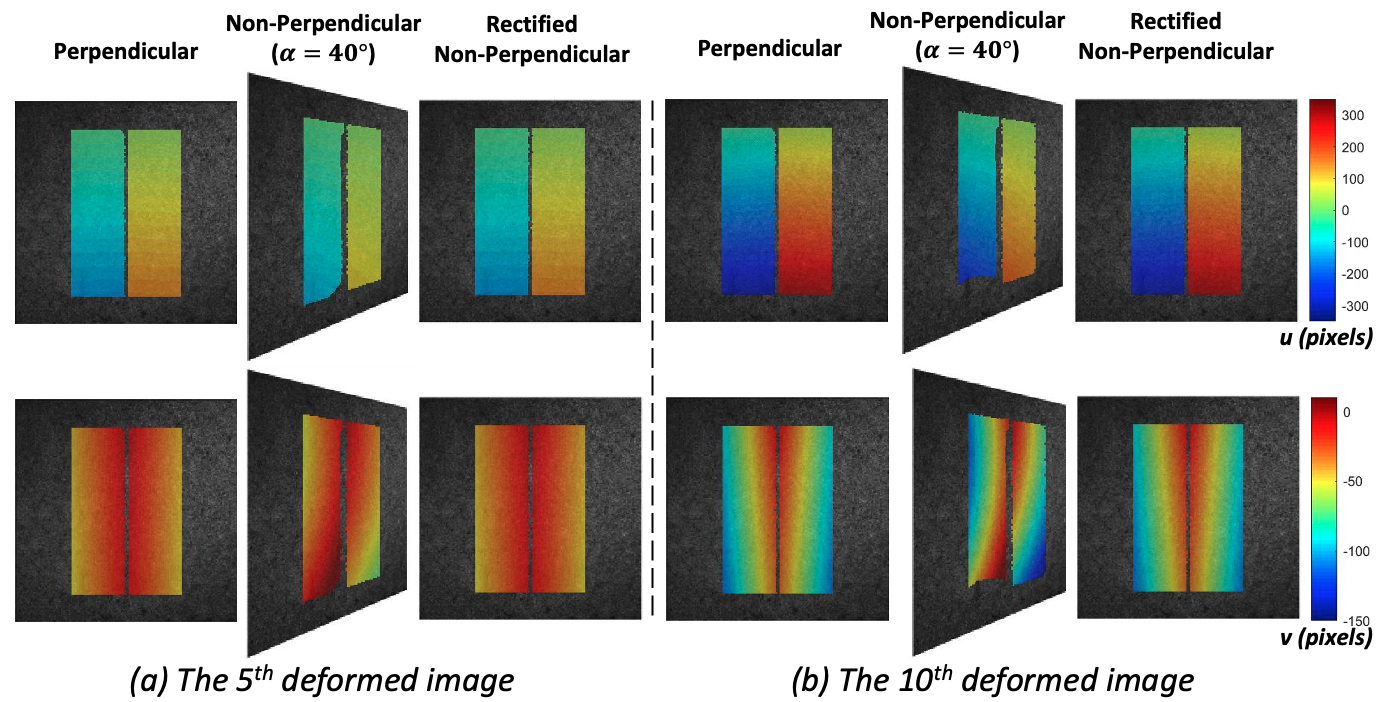}
    \caption{DIC-measured displacement fields on perpendicular, non-perpendicular, and rectified images.}
    \label{fig:num_vis}
\end{figure}

Fig.\ref{fig:num_vis} compares the DIC-measured displacement fields. Due to space limitations, only the fifth and the tenth deformed images of image series Fig.\ref{fig:rotation}(\emph{(f)}) are shown herein. The non-perpendicular camera setting induced significant errors. Meanwhile, the SIFT-aided rectified 2D-DIC method generated visually identical displacement fields as the perpendicular images.

The DIC-measured displacement fields were compared with the ground truth to assess the error quantitatively. The mean absolute error (MAE) was used as the evaluation criterion (Eq.\ref{eqn:mae}). 

\begin{equation}
    \textnormal{MAE} = \frac{\sum_{i=1}^{W} \sum_{j=1}^{H} |u_{ij}'-u_{ij}|}{W\times H}
\label{eqn:mae}
\end{equation}

where $W$ and $H$ are numbers of correlation points in $x$- and $y$-direction, respectively; $u_{ij}'$ is DIC-measured displacement; $u_{ij}$ is actual displacement obtained from Eq.\ref{eqn:displ}.

\begin{figure}[ht!]
    \centering
    \includegraphics[trim=0 0 0 0,clip,width=0.82\textwidth]{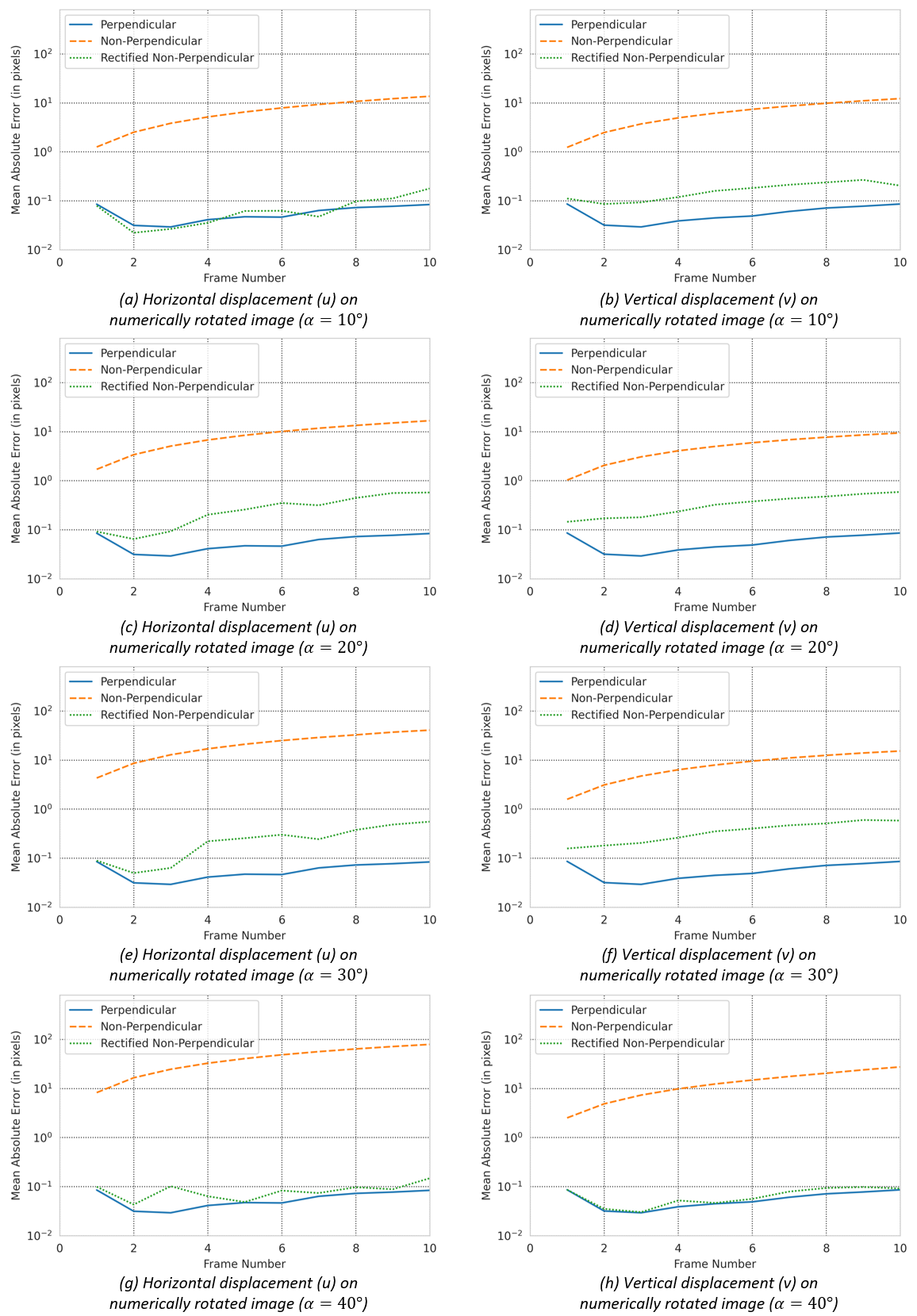}
    \caption{MAE of DIC-measured displacements on the perpendicular, non-perpendicular, and rectified images.}
    \label{fig:res1}
\end{figure}

Fig.\ref{fig:res1} compares the MAE of horizontal ($u$) and vertical ($v$) displacements. The findings are summarized below:
\begin{itemize}
    \item The perpendicular camera setting achieved the smallest error. The MAE was consistently smaller than 0.1 pixels, even under large deformations.  
    \item The highest error was observed on numerically rotated images simulating non-perpendicular camera settings. However, the proposed method significantly reduced the error. The MAE of $u$ and $v$ measured on rectified images remained below 0.6 pixels. This level of accuracy is suitable for typical AC lab testing applications, where the spatial resolution is usually finer than 50$\mu$m/pixel.
    \item The magnitude of deformation and camera rotation angle had a limited impact on the accuracy of SIFT-aided rectified 2D-DIC. The MAE of $u$ and $v$ measured on non-perpendicular images increased as camera rotation angle ($\alpha$) and deformation increased. However, the magnitude of $\alpha$ and deformation had a negligible effect on perpendicular and rectified non-perpendicular images. 
\end{itemize}

\subsection{Experimental Validation}
\label{sec:exp_validation}
\subsubsection{Experimental Setup}
\begin{figure}[ht!]
    \centering
    \includegraphics[trim=0 0 0 0,clip,width=0.98\textwidth]{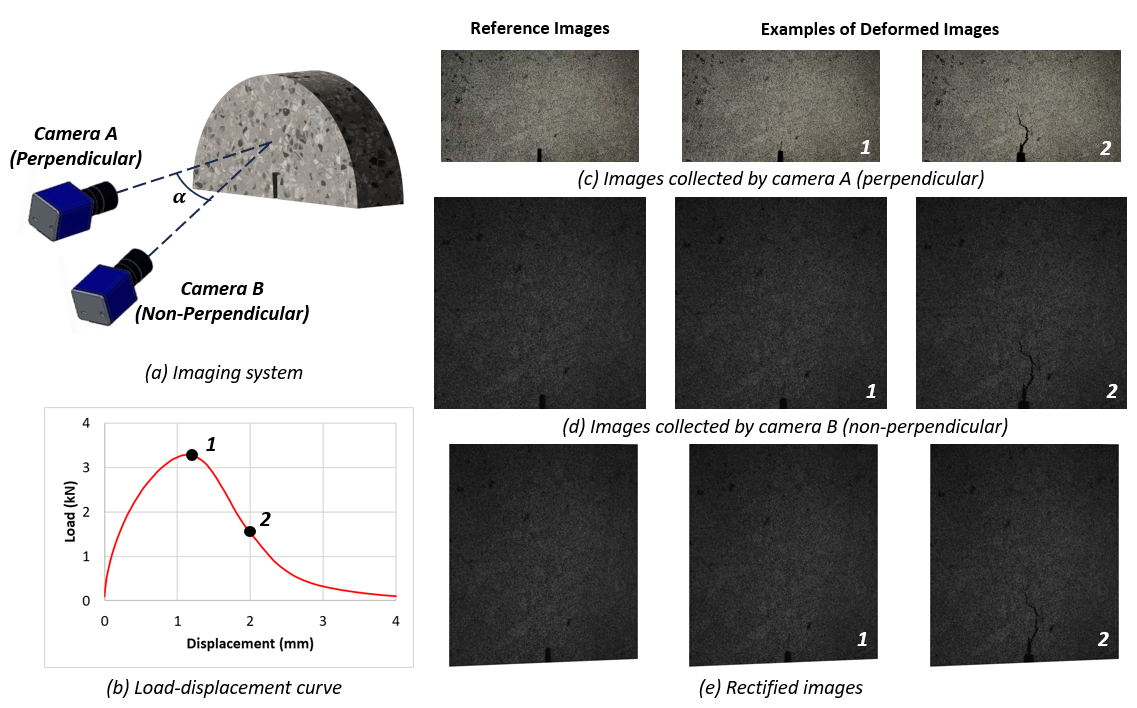}
    \caption{Experimental setup.}
    \label{fig:test_img_example}
\end{figure}

As shown in Fig.\ref{fig:test_img_example}\emph{(a)}, raw images were collected using two synchronized cameras while conducting the I-FIT at 25\textdegree{}C with a 10 mm/min loading rate. A low loading rate was used to minimize the imperfect synchronization effect. A more ductile behavior is expected under a lower loading rate \cite{ozer2016development}. Fig.\ref{fig:test_img_example}\emph{(b)} shows the load-displacement curve. The test specimen's surface was painted with a random black pattern on top of a layer of white paint. Camera A (an iPhone 12 Pro Max's telephoto camera ($1920 \times 1080$ pixels)) was positioned perpendicularly to the specimen surface. Camera B (a Point Grey Gazelle 4.1MP Mono ($2048 \times 2048$ pixels)) was positioned non-perpendicularly. Camera B’s principal axis was rotated $\alpha$ with respect to the z-axis. Before the test, five calibration images were collected in a perpendicular setting using Camera B. 

Fig.\ref{fig:test_img_example}\emph{(c)} and \emph{(d)} show examples of images collected by cameras A and B, respectively. Moreover, the proposed method was followed to rectify images collected by camera B (Fig.\ref{fig:test_img_example}\emph{(e)}). Each image series consists of a reference and 60 deformed images, acquired with an 0.2s time interval. DIC analysis was conducted on the three series of images.

\subsubsection{Results}
Because the actual displacement was unknown, the DIC-measured displacement field on images collected by camera A was used as the baseline. The displacement fields obtained from non-perpendicular and rectified images were compared with the baseline. The mean (Eq.\ref{eqn:mae}) and standard deviation (Eq.\ref{eqn:sdae}) of the absolute error of $u$ and $v$ were used as the evaluation criteria.

\begin{equation}
    \textnormal{SDAE} = \sqrt{\frac{\sum_{i=1}^{W} \sum_{j=1}^{H} (|u_{ij}'-u_{ij}|-\textnormal{MAE}_u)^2}{W\times H}}
\label{eqn:sdae}
\end{equation}

where $W$ and $H$ are numbers of correlation points in $x$- and $y$-direction, respectively; $u_{ij}'$ is measured displacement; $u_{ij}$ is baseline displacement attained from perpendicular images.

\begin{figure}[ht!]
    \centering
    \includegraphics[trim=0 0 0 0,clip,width=0.98\textwidth]{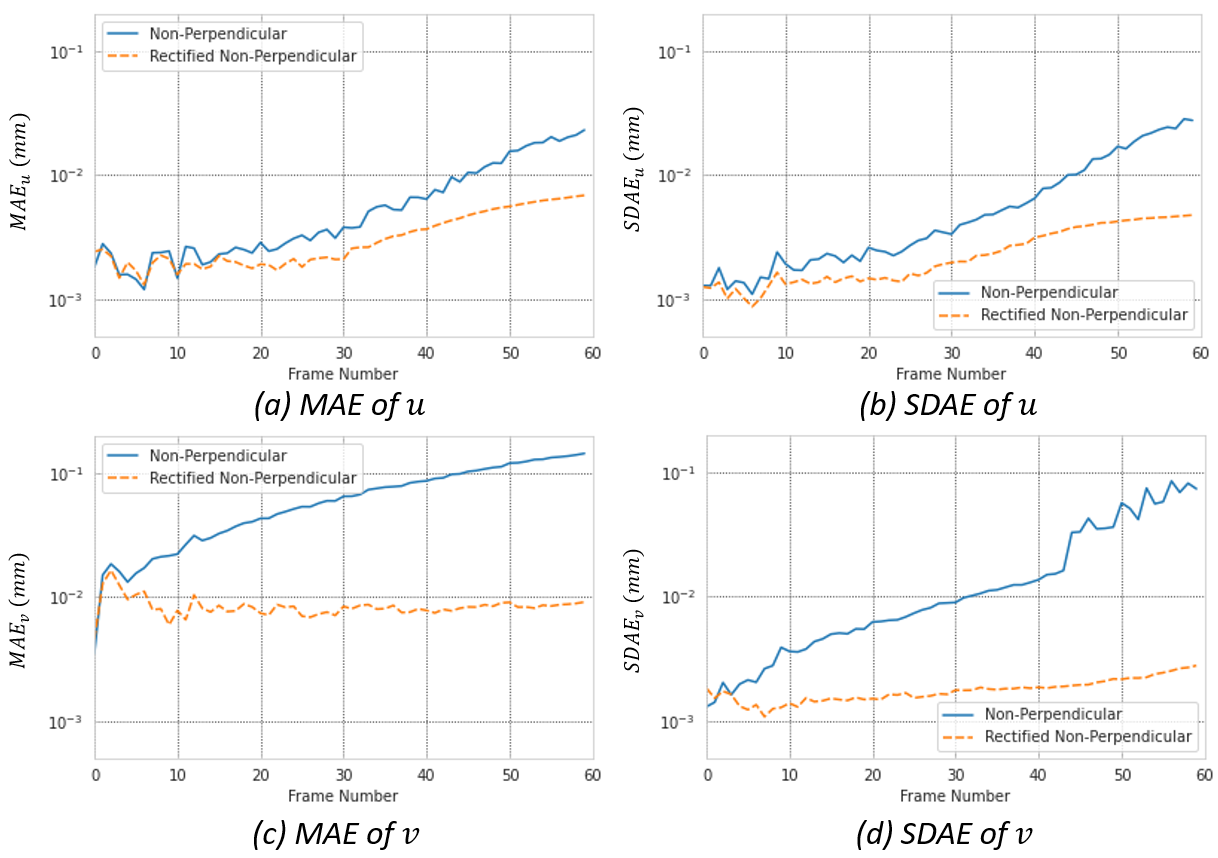}
    \caption{MAE and SDAE of $u$ and $v$ on non-perpendicular images and rectified images.}
    \label{fig:exp_result}
\end{figure}

Fig.\ref{fig:exp_result} shows the MAE and SDAE of $u$ and $v$ on non-perpendicular and rectified images. The findings are summarized below:
\begin{itemize}
    \item The error induced by non-perpendicularity was significant. However, the error was reduced substantially by employing the SIFT-aided rectified 2D-DIC. The MAE and SDAE of both $u$ and $v$ on rectified images were mostly smaller than 0.01 mm. Such accuracy is adequate in an I-FIT application. 
    \item The error on non-perpendicular images increased as the deformation increased. However, such an effect was much less significant for rectified images. 
\end{itemize}

\section{Automated Crack Propagation Measurement with A Non-Perpendicular Camera Alignment}
\citeN{zhu2022auto} recently developed an optical-flow-based deep neural network, CrackPropNet, to measure crack propagation on AC specimen surfaces during cracking tests. The input of CrackPropNet consists of a reference and a deformed image, and it outputs a crack-edge probability map. CrackPropNet locates displacement field discontinuities (i.e., cracks) by matching features at various locations in the reference and deformed image. The network showed promising performance in measuring crack propagation. However, significant errors happened when the camera's principal axis was not perpendicular to the AC specimen surface. The method proposed in this paper can compensate for such errors.

\begin{figure}[ht!]
    \centering
    \includegraphics[trim=0 0 0 0,clip,width=0.98\textwidth]{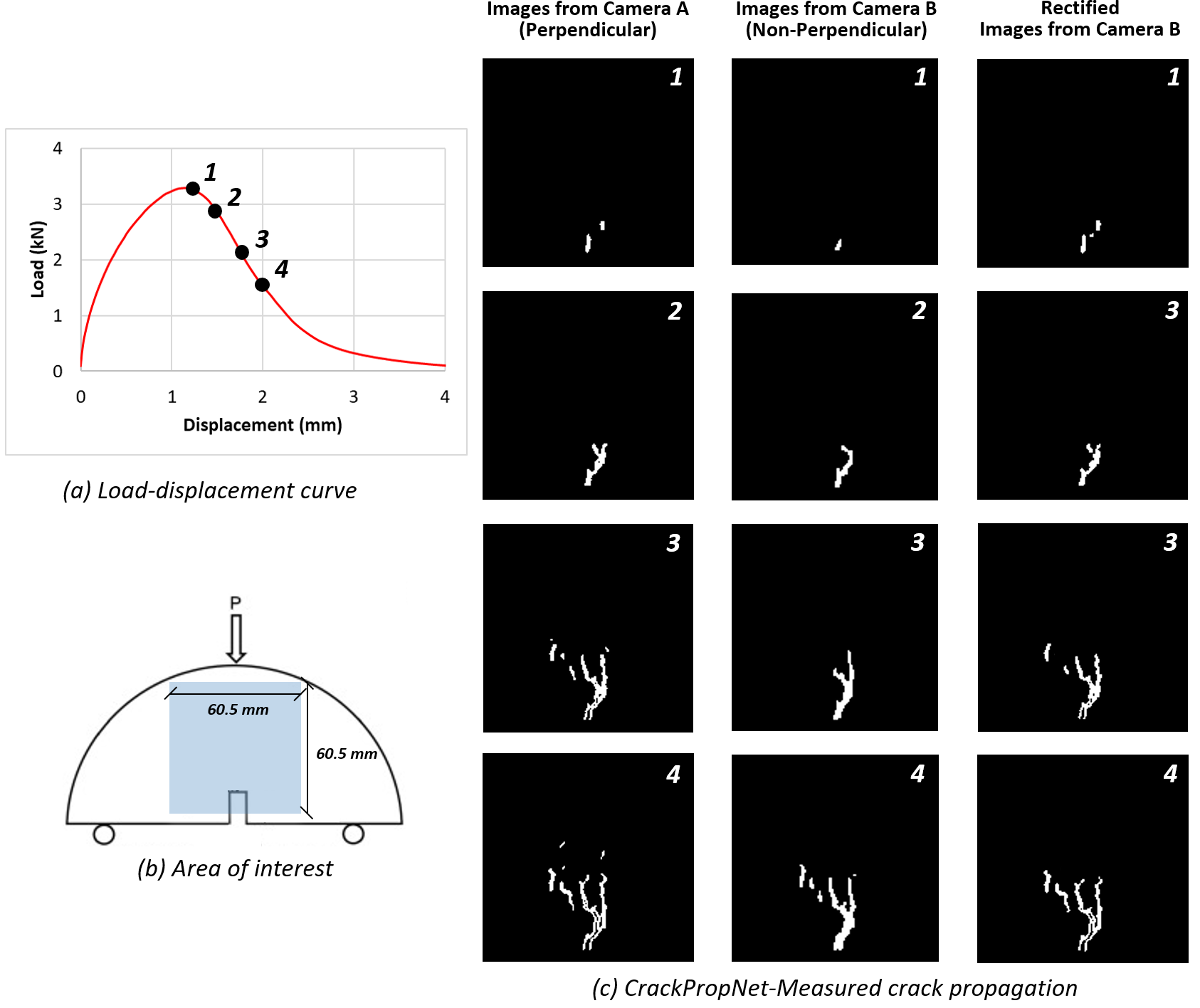}
    \caption{CrackPropNet-measured crack propagation on the perpendicular, non-perpendicular, and rectified non-perpendicular images.}
    \label{fig:exp_propagation}
\end{figure}

As shown in Fig.\ref{fig:exp_propagation}, CrackPropNet was applied to measure crack propagation on the three series of images collected in Fig.\ref{fig:test_img_example}. Because cameras A and B had different fields of view, the crack maps were carefully adjusted to reflect the same square area (Fig.\ref{fig:exp_propagation}\emph{(b)}). Fig.\ref{fig:exp_propagation}\emph{(c)} shows that non-perpendicularity caused non-negligible measurement errors. Due to the rotation of camera B's principal axis around the z-axis, the cracks experienced compression in the horizontal direction, leading to the loss of crack shape details. It is expected that CrackPropNet would fail more drastically if the camera rotation angle is larger. Conversely, the rectification procedure effectively restored most crack shape details, resulting in crack measurements similar to those obtained from the perpendicular images, with only minor differences.

To quantitatively evaluate the accuracy of the proposed method, an experiment was conducted on the testing dataset developed by \citeN{zhu2022auto}. The dataset comprises 188 images collected while conducting the I-FIT on various AC mixes. It covers the diversified AC cracking behavior. Cameras were positioned perpendicular to I-FIT specimen surfaces to collect these images. 

\begin{figure}[ht!]
    \centering
    \includegraphics[trim=0 0 0 0,clip,width=0.98\textwidth]{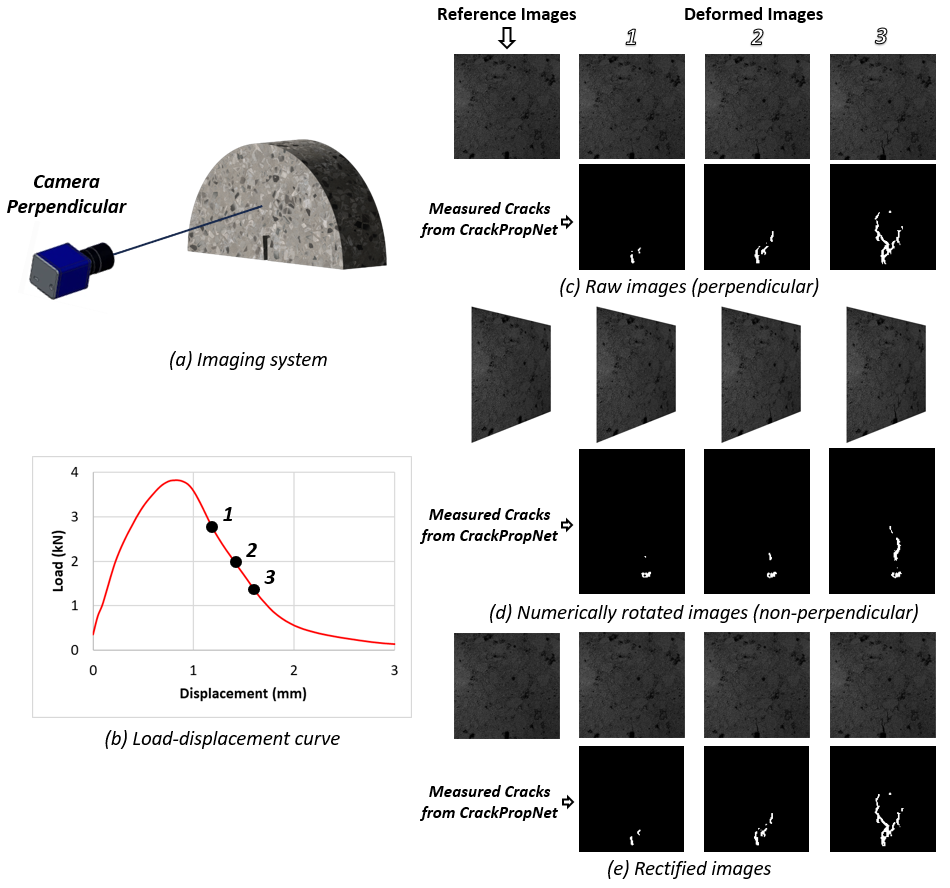}
    \caption{CrackPropNet-measured crack propagation on raw, numerically rotated, and rectified images.}
    \label{fig:net_result}
\end{figure}

For example, Fig.\ref{fig:net_result}\emph{(c)} shows a reference image and its related deformed images, which correspond to points 1, 2, and 3 in the load-displacement curve. To simulate a non-perpendicular camera setting, raw images were numerically rotated $40^\circ$ around the z-axis (Fig. \ref{fig:net_result}\emph{(d)}). These numerically rotated images were then rectified using the SIFT-aided rectification procedure (Fig. \ref{fig:net_result}\emph{(e)}). The raw, numerically rotated, and rectified images were subsequently processed by CrackPropNet for crack propagation measurement. Notably, Fig. \ref{fig:net_result}\emph{(d)} demonstrates that non-perpendicularity introduced significant errors, with cracks measured from numerically rotated images missing crucial details. Conversely, Fig. \ref{fig:net_result}\emph{(e)} showcases the effectiveness of the proposed method, which successfully recovered crack shape details and produced nearly identical results to those obtained from perpendicular images.

The above procedure was repeated for the entire dataset. The measured cracks were compared with the ground-truth crack labels provided by \citeN{zhu2022auto}. F-1 ($\frac{2 \cdot Precision \cdot Recall}{Precision + Recall}$) was used as the evaluation criterion, where a higher F-1 indicates a more accurate measurement. To generate the crack map based on the CrackPropNet's probability map output, two commonly used strategies were employed: optimal dataset scale (ODS) and optimal image scale (OIS). The former employs a fixed threshold for all images, while the latter finds an optimal threshold for each image. The CrackPropNet achieved an ODS F-1 of 0.748 and an OIS F-1 of 0.778 on rectified images, while it achieved an ODS F-1 of 0.755 and an OIS F-1 of 0.781 on perpendicular images.


It is safe to conclude that the proposed method can efficiently compensate for crack propagation measurement errors caused by non-perpendicularity.



\section{Summary and Conclusions}
This paper proposes the SIFT-aided rectification method to compensate for 2D-DIC measurement errors caused by non-perpendicular camera alignment. The proposed method is simple and easy-to-use because no additional equipment is needed. A rigorous experimental program was conducted to demonstrate its accuracy and reliability.

First, a theoretical error analysis was conducted to quantify the effects of a non-perpendicular camera alignment on displacement measurement. Under the pinhole camera model, focal length ($f$), pinhole-object distance ($S$), camera rotation angle ($\theta$), displacement ($\Delta x$ \& $\Delta y$), and pixel location ($x_A$ \& $y_A$) significantly affect the absolute error of projected displacement on the image plane.

Second, the proposed method was validated numerically using synthetic images. The MAE of horizontal and vertical displacements on rectified images was consistently smaller than 0.6 pixels, even under considerable camera rotation (up to 40$^{\circ}$) and large deformation. Such accuracy is adequate in typical AC lab testing applications, where spatial resolution is normally finer than 50$\mu$m/pixel. 

Third, the proposed method was experimentally verified. Raw images were collected using two synchronized cameras while conducting the I-FIT. DIC-measured displacement field on images collected under a perpendicular setting was used as the baseline. The SIFT-aided rectified 2D-DIC efficiently compensated for errors induced by non-perpendicular camera settings. The MAE and SDAE of horizontal and vertical displacement were predominantly below 0.01 mm.

Fourth, the SIFT-aided rectification method was validated as a viable approach to support CrackPropNet in measuring crack propagation under non-perpendicular camera settings. Both qualitative and quantitative evaluations confirmed that the rectification procedure effectively restored most crack shape details from non-perpendicular images, leading to crack measurements that closely aligned with those obtained from perpendicular images.

The following conclusions are drawn from this study:
\begin{itemize}
    \item Non-perpendicular camera alignment significantly impacts 2D-DIC measurement accuracy during asphalt concrete testing.
    \item The proposed SIFT-Aided Rectified 2D-DIC method significantly reduces measurement error caused by a non-perpendicular camera alignment, even with substantial camera rotation and relatively large deformation.
    \item The SIFT-aided rectification method could be applied to assist CrackPropNet in measuring crack propagation under non-perpendicular camera settings.
\end{itemize}

\section{Limitations and Recommendations}
The followings are the limitations of this study and relative suggestions for future research:
\begin{itemize}
    \item The proposed method is applicable to planar specimen surfaces. A stereo-DIC is needed for non-planar objects, such as a cylindrical dynamic modulus specimen.
    \item Other key point detection and matching algorithms, such as speeded-up robust features (SURF) and deep-learning-based SuperGlue may be considered in the future. An end-to-end deep neural network insensitive to non-perpendicular camera alignment may be developed by stacking SuperGlue and CrackPropNet.
\end{itemize}

\section{Data Availability Statement}
All data, models, or code supporting this study's findings are available from the corresponding author upon reasonable request.

\section{Acknowledgment}
The authors thank the research engineers and students at the Illinois Center for Transportation: Greg Renshaw, Uthman Mohamed Ali, and Jose Julian Rivera-Perez. The authors hold responsibility for the accuracy and factual information presented in the contents.

%
%
\bibliography{ascexmpl-new}

\end{document}